\documentclass[letterpaper, 10 pt, conference]{ieeeconf}
\IEEEoverridecommandlockouts
\usepackage{cite}
\usepackage{amsmath,amssymb,amsfonts}
\usepackage{algorithm}
\usepackage{algorithmic}
\usepackage{graphicx}
\usepackage{textcomp}
\usepackage{xcolor}
\usepackage{makecell}
 
\def\BibTeX{{\rm B\kern-.05em{\sc i\kern-.025em b}\kern-.08em
    T\kern-.1667em\lower.7ex\hbox{E}\kern-.125emX}}
\begin{document}

\title{ A Framework for eVTOL Performance Evaluation in Urban Air Mobility Realm 
\thanks{The authors would like to thank the National Institute of Aerospace's Langley Distinguished Professor Program and the NASA University Leadership Initiative (ULI).}
}

\author{Mrinmoy Sarkar, Xuyang Yan, Abenezer Girma, and Abdollah Homaifar\\
North Carolina A\&T State University,
Greensboro, North Carolina, 27411\\
Emails: msarkar@aggies.ncat.edu, xyan@aggies.ncat.edu, aggirma@aggies.ncat.edu, homaifar@ncat.edu}

\maketitle

\begin{abstract}
In this paper, we developed a generalized simulation framework for the evaluation of electric vertical takeoff and landing vehicles (eVTOLs) in the context of Unmanned Aircraft Systems (UAS) Traffic Management (UTM) and under the concept of Urban Air Mobility (UAM). Unlike most existing studies, the proposed framework combines the utilization of UTM and eVTOLs to develop a realistic UAM testing platform. For this purpose, we first enhanced an existing UTM simulator to simulate the real-world UAM environment. Then, instead of using a simplified eVOTL model, a realistic eVTOL design tool, namely SUAVE, is employed and an dilation sub-module is introduced to bridge the gap between the UTM simulator and SUAVE eVTOL performance evaluation tool to elaborate the complete mission profile. Based on the developed simulation framework, experiments are conducted and the results are presented to analyze the performance of eVTOLs in the UAM environment. 
\end{abstract}


\section{Introduction} \label{introduction}
To address ground traffic congestion, NASA initiated the ``Urban Air Mobility" concept to utilize the three-dimensional airspace to accommodate the heavy demand for cargo deliveries as well as passenger transportation in urban areas. The \textbf{e}lectric \textbf{V}ertical \textbf{T}akeoff and \textbf{L}anding (eVTOL)\footnote{``AHS International Leads Transformative Vertical Flight Initiative". evtol.news. Retrieved 2020-09-23.} aircraft is proposed as a promising solution to implement UAM in the future. Under this direction, different types of eVTOL have been extensively investigated by researchers from the industry such as Uber, Joby Aviation and Boeing. It is envisioned that eVTOLs will significantly reduce the heavy traffic congestion during peak times and improve the efficiency of urban traffic networks.\\
\indent The safety and reliability of UAM have motivated extensive research works for both design and evaluation of UAM and eVTOLs. In \cite{sunil2016influence,zhu2016low,peinecke2017deconflicting,jang2017concepts,joulia2016towards,bulusu2018throughput}, the authors primarily focused on developing different alternatives for the management of UAM with respect to different objectives however, no systematic approach is available to generate all possible alternatives. Also, most UTM alternatives are evaluated in different simulation interfaces and no common platform is yet developed for the evaluation of these alternatives. Considering these challenges, a preliminary framework with metrics and a simulation interface was developed to compare and evaluate different alternatives for UAM in \cite{ramee2021development}. However, it simplified eVTOL operations from 3-D to 2-D space and ignored the physical configuration of the eVTOL's shape, weight, and battery efficiency. Moreover, the preliminary framework did not consider the practical mission profiles of eVTOL, and thus cannot effectively model their behaviors in real-world scenarios.\\
\indent The open-source conceptual design tool, namely SUAVE \cite{lukaczyk2015suave}, is capable of modeling the physical properties of eVTOLs and incorporating the practical mission profile of eVTOLs. In \cite{clarke2019strategies}, SUAVE is used to optimize the design of eVTOLs by identifying the relationship between vehicle configurations and performance without considering the interactions with other eVTOLs in UAM. However, in real-world scenarios, the operational environment of eVTOLs is dense and interaction from other eVTOLs must be considered. In summary, several limitations of the existing studies of UAM and eVOTLs are as follows: 
\begin{itemize}
\item According to authors' knowledge, no prior work has developed a framework for the evaluation of eVTOLs in a fully-realized UAM environment.
\item The trade-off between fidelity and scale is not considered in the existing UTM simulation frameworks; High fidelity takes a longer time with  poor scalability while low fidelity may ignore infeasible missions with better scalability. 
\item Existing testing and design tools of eVTOLs rarely consider the complexity of the operational environment in UAM.
\end{itemize}

\indent In light of these limitations, we develop an effective framework to evaluate the performance of eVTOLs in UAM. The proposed framework can not only simulate the interactions of eVTOLs in a highly congested UAM network but also mimic the realistic properties of eVTOLs. With a low-fidelity UTM simulator, the proposed framework can provide detailed analysis to identify and explain infeasible mission profiles, which partially consider the trade-off effect between fidelity and scale. The contributions of this work are two-fold: First, we integrate the preliminary framework from \cite{ramee2021development} with the SUAVE tool to develop a more comprehensive evaluation framework for eVTOLs. Second, we conduct simulations using the developed framework and present the simulation results to analyze the performance of eVTOLs in a UAM network.\\
\indent The remainder of this paper is organized as follows: Section \ref{literature_review} provides a review of different research studies in UTM and eVTOL testing. The details of the proposed framework are described in Section \ref{methodology}. Section \ref{results} presents experimental studies using the proposed framework. The advantages of the proposed framework are summarized in Section \ref{discussion}. Finally, concluding remarks and future work are outlined in Section \ref{conclusion}.

\section{Literature Review} \label{literature_review}
The demand for a new air traffic management system to coordinate increasingly dense low-altitude flights has recently attracted substantial research \cite{sunil2016influence, zhu2016low, peinecke2017deconflicting, jang2017concepts, bulusu2018throughput}. Most of these research studies propose different air-traffic management methods and testing tools to evaluate the effectiveness of aerial vehicle (such as eVTOL) designs under different flight profiles and objectives. 
In this section, we review those studies and identify gaps to motivate a more realistic aerial vehicle testing and evaluation framework.

\textbf{UTM Simulation Framework for UAM:}
A systematic approach to generate all viable architectures, as required by a thorough decision-making process, is still under development. Some architectures are still at the Concept of Operations (ConOps) stage and have not been thoroughly evaluated. Although some other architectures have been tested in simulation environments, each approach is tested with a distinct simulator and some simulators are not publicly available. This makes the evaluation and comparison of the architectures difficult. For instance, architectures such as Full Mix, Layers, Zones, and Tubes, are developed using the TMX simulator \cite{chambers2011reforms}, while the Iowa UTM \cite{zhu2016low} and Altiscope \cite{effectiveness2018} architectures utilize the Custom 2D simulator. Moreover, DLR Delivery Network \cite{peinecke2017deconflicting}, Linköping distributed \cite{sedov2018centralized} and Linköping centralized \cite{sedov2018centralized} architectures are built on the Custom 3D simulator. 

To address these gaps and design a better decision-making process for UTM, recent work in \cite{ramee2021development} has proposed an open-source simulation framework that can generate various alternatives for different UTM subsystems. In addition, the framework can compare alternatives based on safety, efficiency, and capacity. The framework consists of four elements: system decomposition, alternative generation, comparison metric establishment, and alternative evaluations. To create general alternatives for the conceptual design, the author decomposed the UTM into four subsystems: airspace structure, access control, preflight planning, and collision avoidance. 

The author used a custom-built open-source 2D agent-based simulation framework 
to generate and evaluate alternatives. However, the 2D simulation framework fails to capture important factors of a real-world environment, including the agent's altitude and wind condition. Additionally, the agents are modeled as 2D points in the simulator, and the vehicle's sizes, shapes, and dynamics are ignored, which hinders the practicality of the simulator for modeling real-world scenarios. In the real-world environment, the physical design of the vehicles plays an important role in studying the effect of aerodynamics, airspace capacity, weather, and other factors. 

\begin{figure}[th]
    \centering
    \includegraphics[width=0.45\textwidth]{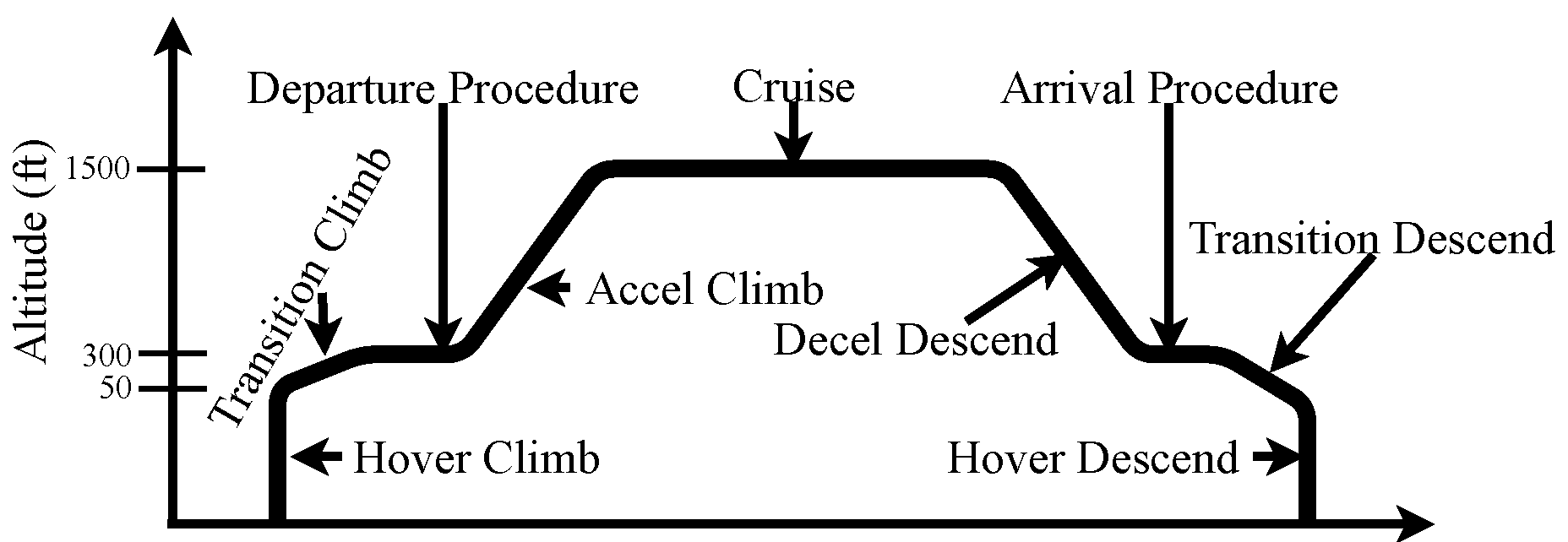}
    \caption{A generalized eVTOL flight profile \cite{UBAR_mission_spec}. }
    
    \label{fig:general_mission_profile}
\end{figure}

Moreover, it is assumed that the vehicles stay at the same altitude throughout the flight and a simple 2D approach is used for the modeling. Accordingly, the take-off and the landing procedures are completely ignored by the study. This simplification reduces the framework's complexity by omitting one of the essential components of the overall flight process where there are numerous uncertainties and challenges. Once the agents are granted access to the airspace, each agent optimizes their own trajectory using either the decoupled method, safe interval path planning, or local path planning. The collision avoidance system is implemented using a reactive, decentralized strategy called Modified Voltage Potential (MVP) \cite{eby1994self,hoekstra2016bluesky}. Finally, the UTM's performance is evaluated based on the established objectives. 
\textbf{eVTOL Design and Performance Analysis Tool} : In \cite{clarke2019strategies}, a method is proposed for testing different types of eVTOL designs in accomplishing a given mission profile. It uses SUAVE \cite{lukaczyk2015suave} to achieve a realistic design of various types of aircraft configurations with varying aerodynamic fidelity analyses. 
After designing and optimizing the eVTOLs in SUAVE, each vehicle is tested with a static mission profile. As shown in Figure \ref{fig:general_mission_profile}, the mission profile starts with an initial take-off and it is followed by the ascent, cruise, descent, and reserve hover/loiter. Based on a fixed mission profile the performance of different eVTOL designs is evaluated.


Despite the fact that the study in \cite{clarke2019strategies} considers testing of different realistic eVTOL designs, it fails to study the effect of different dynamic mission profiles on the performance of the eVTOL. In a real-world scenario, the eVTOL should be robust enough to follow different mission profiles generated based on various environmental conditions and circumstances. However, \cite{clarke2019strategies} considered a fixed mission profile where the eVTOL follows a pre-defined mission that is fixed in terms of time and altitude. 
Under congested low-altitude UAM conditions, the speed of the eVTOL can be affected by different internal and environmental factors, such as eVTOL payloads, weather conditions, and collision avoidance system. The author also assumes a collision-free path, which is unrealistic. 

\begin{table}[!h]
\small
\unboldmath
\caption{Comparison of studies in literature.}
\begin{tabular}{l|l|l|l}
\hline
\makecell{Features} & \makecell{UTM \\ Simulator\cite{ramee2021development}}  & \makecell{SUAVE \cite{clarke2019strategies}} & \makecell{Proposed}  \\ \hline
  
  \makecell{Realistic UAM \\ simulation \\ environment} & \checkmark   & $\times$ & \checkmark   \\\cline{1-1}
    
  
   \makecell{Real-eVTOL \\ dynamics \\ and configuration}   & $\times$   &\checkmark & \checkmark  \\\cline{1-1}
    
    
    \makecell{Dynamic mission \\profile}
      & \checkmark   & $\times$ & \checkmark \\\cline{1-1}
        
    \makecell{Evaluate each \\ segment of the \\mission profile} & $\times$    & \checkmark & \checkmark \\\cline{1-1}
    
    \makecell{Can test \\ a novel-realistic \\ aircraft design} & $\times$    & \checkmark & \checkmark \\
        
  \hline
\end{tabular}
\label{tab:simulator_features}
\end{table}

In this paper, we aim to address these gaps. We propose an eVTOL performance testing and evaluation framework that leverages the advantages of the realistic UTM simulator from \cite{ramee2021development, eby1994self, lukaczyk2015suave} and a practical eVTOL design and flight profiling tool from \cite{clarke2019strategies, lukaczyk2015suave}. \textcolor{black}{In Table \ref{tab:simulator_features}, we summarized the features of the existing UTM simulation framework and eVTOL performance analysis tool with the proposed simulation framework.}

\begin{figure*}[!th]
    \centering
    \includegraphics[scale=0.35]{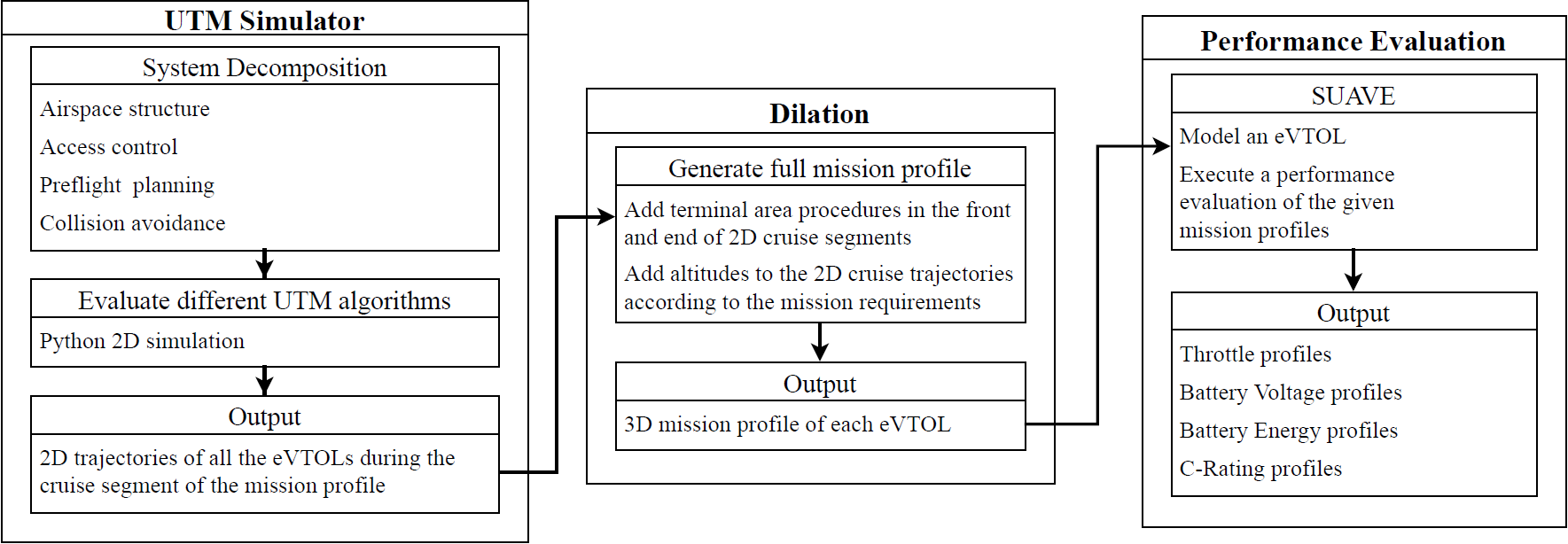}
    \caption{The architecture of the integrated simulation framework.}
    \label{fig:system_architecture}
\end{figure*}

\section{Methodology}\label{methodology}
As described in Sections \ref{introduction} and \ref{literature_review}, the existing studies provide UTM and eVTOL performance evaluation independently. Nevertheless, the concept of UAM can be only fully realized when UTM and eVTOL performance evaluations are integrated. In this paper, we combined these two different performance evaluation schemes and developed a new simulation framework. With this new framework, we can analyze the limitations and strengths of any UTM algorithm or realistic eVTOL model within a UAM environment. The system architecture of the proposed integrated simulation framework is shown in Figure \ref{fig:system_architecture}. It consists of three sub-modules: (1) UTM Simulator, (2) Dilation, and (3) Performance Evaluation. Details of each sub-module are described in the following sub-sections.

\subsection{UTM Simulator}
The UTM Simulator is developed by adapting the open source implementation of the UTM simulation framework from \cite{ramee2021development}. It uses off-the-shelf implementations of the airspace structure, access control, preflight planning, and collision avoidance algorithms. We modify the visualization tool of the original implementation to make it realistic and extract the entire trajectory of each agent/eVTOL. This new feature provides more details about the behaviors of eVTOLs at different time intervals during the overall mission. \textcolor{black}{Since the original implementation of the UTM simulator in \cite{ramee2021development} ignores vertiport \footnote{A type of airport for aircrafts which take off and land vertically.} terminal area procedures (Standard Instrument Departures (SIDs), Standard Instrument Arrivals
(STARs) and Approaches etc.) by focusing only on the cruise segment of the entire flight profile, we extract only the 2D trajectory of the cruise segment in our UTM simulator sub-module.} Thus, the output of the UTM simulator in our framework is a set of 2D trajectories of all the eVTOLs in the simulation, which enter the airspace at a specific time and finish an entire flight. Each entry of the 2D trajectory is composed of five elements: x-coordinate, y-coordinate, $v_x$-velocity, $v_y$-velocity, and time-stamp.

\subsection{Dilation}
We elaborate the cruise segment with other required segments such as takeoff, transition, climb, descent and land in the dilation sub-module. Accordingly, the output of the dilation algorithm is a full mission profile of each eVTOL. Moreover, we convert the trajectories into 3D trajectories by incorporating altitudes for all the entries in the 2D trajectories. The algorithmic description of the dilation sub-module is shown in algorithm \ref{algo:ecapsul}. \textcolor{black}{Though real-world implementation of vertiport terminal area procedures requires further investigation, the Hover Climb, Transition Climb, Departure Terminal area Procedures, Arrival Terminal area Procedures, Transition Descend and Hover Descend segments in the dilation algorithm can be inferred as vertiport terminal area procedures.}

\begin{algorithm}[h] 
\caption{Creation of Full Mission profiles for each eVTOL} 
\label{algo:ecapsul} 
\begin{algorithmic} 
    \REQUIRE 2D Trajectories generated from UTM Simulator
    \ENSURE 3D Trajectories with initial and ending segments of the mission profile
    \STATE $\zeta \Leftarrow \emptyset$
    \STATE $N \Leftarrow \text{Number of 2D Trajectories}$
    \FOR{$i=1$ \TO $N$ } 
    \STATE $\tau \Leftarrow \emptyset$
    \STATE Append Hover Climb Segment to $\tau$
    \STATE Append Transition Climb Segment to $\tau$
    \STATE Append Departure Terminal area Procedures Segment to $\tau$
    \STATE Append Accelerated Climb Segment to $\tau$
    \STATE Insert a constant altitude to each waypoint of the $i^{th}$ 2D Trajectory
    \STATE Append the $i^{th}$ 3D Trajectory as the Cruise Segment to $\tau$
    \STATE Append Decelerated Descend Segment to $\tau$
    \STATE Append Arrival Terminal area Procedures Segment to $\tau$
    \STATE Append Transition Descend Segment to $\tau$
    \STATE Append Hover Descend Segment to $\tau$
    \STATE Append $\tau$ to $\zeta$
    \ENDFOR
    \RETURN $\zeta$
\end{algorithmic}
\end{algorithm}

\subsection{Performance Evaluation}
To evaluate the performance of each eVTOL in the UAM environment, we use the open-source software tool SUAVE \cite{SUAVEGit, SUAVE2017}. SUAVE is a set of tools to design and optimize conceptual novel aircraft. As an example, we used an eVTOL model developed in SUAVE. The elaborated mission profiles generated by the dilation sub-module are used in SUAVE to evaluate the performance of the eVTOL model. There are several evaluation criteria, however, four built-in metrics, including Throttle profile, Battery energy profile, Battery voltage profile, and C-Rating profile, are employed to evaluate the performance of the eVTOL for all the mission profiles. The metrics are described as follows: 

\paragraph{Throttle profile}
In general, the throttle controls the vertical motion of an eVTOL and this measurement is directly proportional to the thrust generated from its motors. From the throttle profile, the amount of thrust contribution from each motor during different segments of the mission profile are obtained.

\paragraph{Battery Voltage profile}
Using the battery voltage profile, we can observe the decreasing trend of battery voltage for the eVTOL as the mission progresses.

\paragraph{Battery Energy profile}:
This profile shows the battery energy consumption by the eVTOL along the mission profile. It is an important metric for eVTOL performance measurement because battery energy consumption is directly related to the range of the mission that can be achieved by the eVTOL.

\paragraph{C-Rating profile}
The C-Rating profile shows the battery discharge rate of the eVTOL for the given mission profile.  

\section{Results} \label{results}
We present our results from the integrated eVTOL performance measurement framework and those details are discussed below.

\begin{figure}[th]
    \centering
    \includegraphics[scale=0.07]{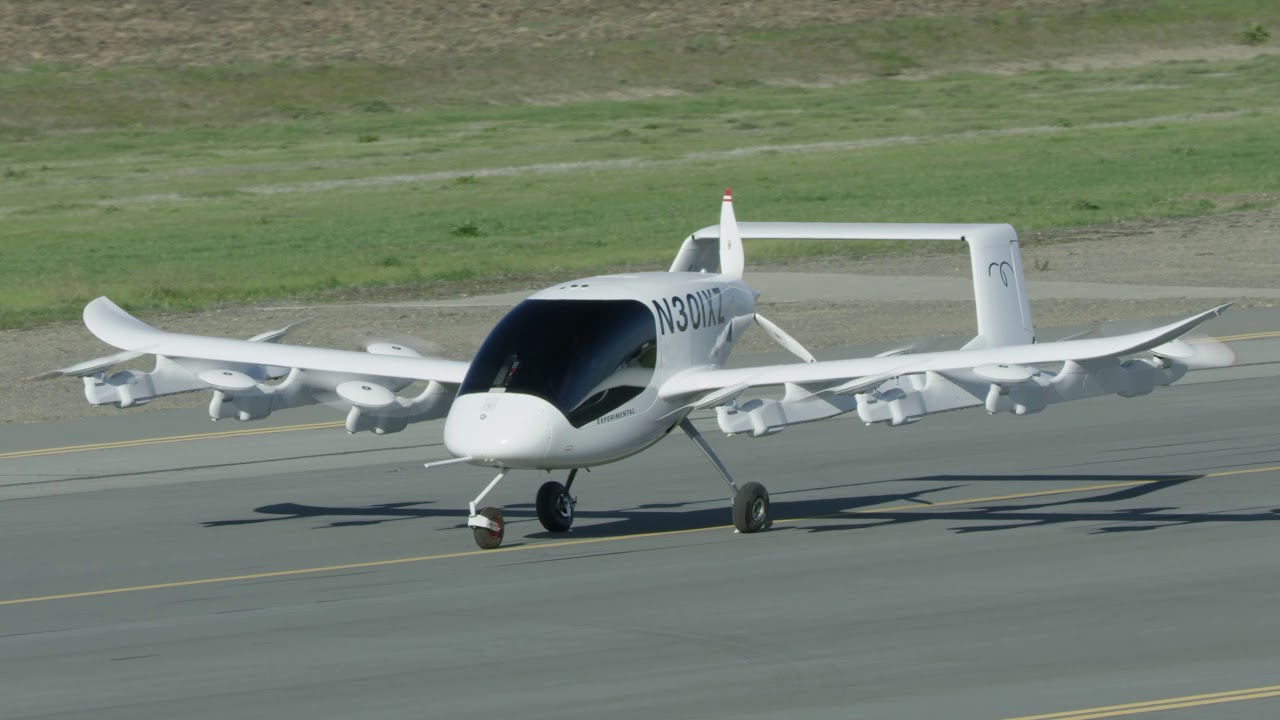}
    \includegraphics[scale=.6]{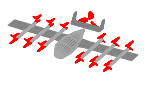}
    \caption{The kitty Hawk Cora eVTOL (left) and SUAVE-Open VSP generated eVTOL model (right).}
    \label{fig:evtol_model}
\end{figure}

For the UTM simulator, a set of algorithms are selected  \cite{ramee2021development} and Table \ref{tab:utm_sim_parameters} summarizes all the parameters. \textcolor{black}{We use ``free airspace structure," meaning all eVTOL can fly their preferred path to their destination and ``free access control" which allows an eVTOL to take-off if there is no immediate conflict. In this simulation study, no preflight planning is used but a reactive decentralized strategy known as Modified Voltage Potential (MVP) algorithm is used for collision avoidance.} The eVTOL model used in our experiment is developed based on the geometry of the Kitty Hawk Cora eVTOL prototype. This eVTOL configuration is also known as a lift+cruise configuration. The original Kitty hawk Cora eVTOL and the generated eVTOL from SUAVE tool are shown in Figure \ref{fig:evtol_model}. Some of the high-level parameters of the eVTOL are listed in Table \ref{tab:eVtol_param}. The complete list of parameters of the eVTOL model can be found in the SUAVE GitHub repository \footnote{\label{evtol_param} https://github.com/suavecode/SUAVE/\\blob/develop/regression/scripts/Vehicles/Stopped\_Rotor.py}.

\begin{table}[h]
    \centering
    \caption{List of parameters used for the UTM simulator}
    \label{tab:utm_sim_parameters}
    \begin{tabular}{c|c} \hline
         Parameter name & Value \\ \hline 
         Operational area & $50\times50\text{ } km^2$ \\ 
         Minimum separation & $500\text{ }m$ \\ 
         Sensing radius & $5000\text{ }m$ \\ 
         Max speed of the eVTOL & $100.662\text{ }mph$\\ \hline 
    \end{tabular}
\end{table}

\begin{table}[th]
    \centering
    \caption{A set of high-level parameters of the considered eVTOL model\textsuperscript{\ref{evtol_param}}.}
    \label{tab:eVtol_param}
    \begin{tabular}{c|c} \hline
        Parameter name &  Value \\ \hline
         Max Takeoff mass & $2450\text{} lbs$ \\
         Max Payload mass & $200 \text{}lbs$ \\
         Total reference area & $10.76$ $m^2$ \\
         Number of lift motor & $12$ \\
         Number of cruise motor & $1$  \\
         Battery type & Lithium Ion \\
         Battery max voltage & $500\text{} volt$ \\
         Battery energy specific density & $300\text{} Wh/kg$ \\
         $v_{stall}$ & $84.28\text{} mph$\\
         Speed & $111.847\text{} mph$\\ \hline
    \end{tabular}
\end{table}

From the UTM simulator module we generated $262$ 2D trajectories using the simulation parameters from Table \ref{tab:utm_sim_parameters}. We then used the dilation algorithm \ref{algo:ecapsul} to convert the 2D trajectories into full mission profiles as shown in Figure \ref{fig:general_mission_profile}. During the dilation procedure, we referred to the required specification of a UAM mission for eVTOL from Uber Elevate \cite{UBAR_mission_spec, holden2016fast}, which are described in Table \ref{tab:mission_spec}. However, we kept the altitude as Table \ref{tab:mission_spec}, but chose other parameters such as vertical speed and horizontal speed randomly from a bound of $[\mu-\Delta,\mu+\Delta]$, where $\mu$ are the values showed in Table \ref{tab:mission_spec}. This implementation is inspired by the different environmental conditions (both geographical and weather) at different vertiport locations.

\begin{figure}
    \centering
    \includegraphics[scale=0.52]{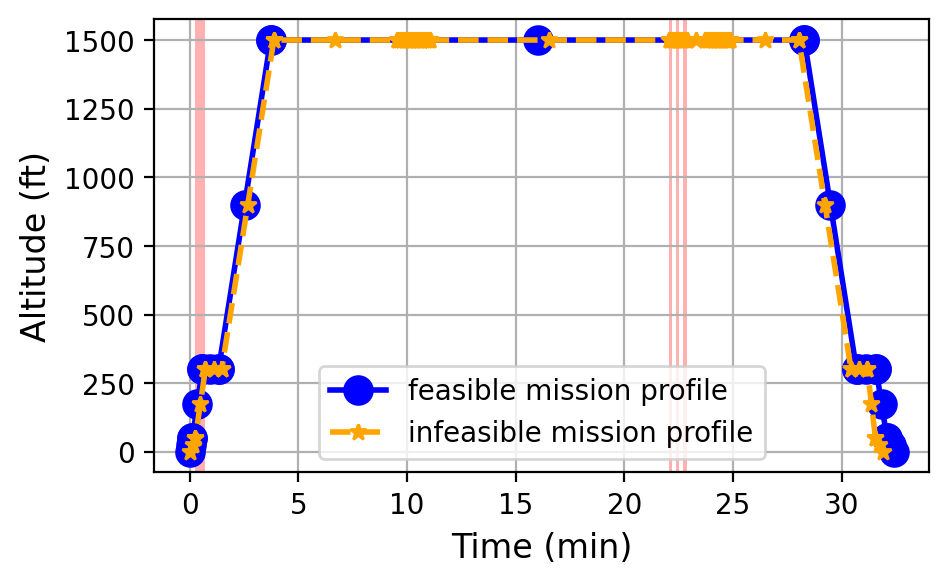}
    \caption{A sample feasible and infeasible mission profile. The red vertical bars highlight the segments of the infeasible mission profile where SUAVE eVTOL was unable to execute. }
    \label{fig:mission_profile}
\end{figure}

\begin{figure}
    \centering
    \includegraphics[scale=0.52]{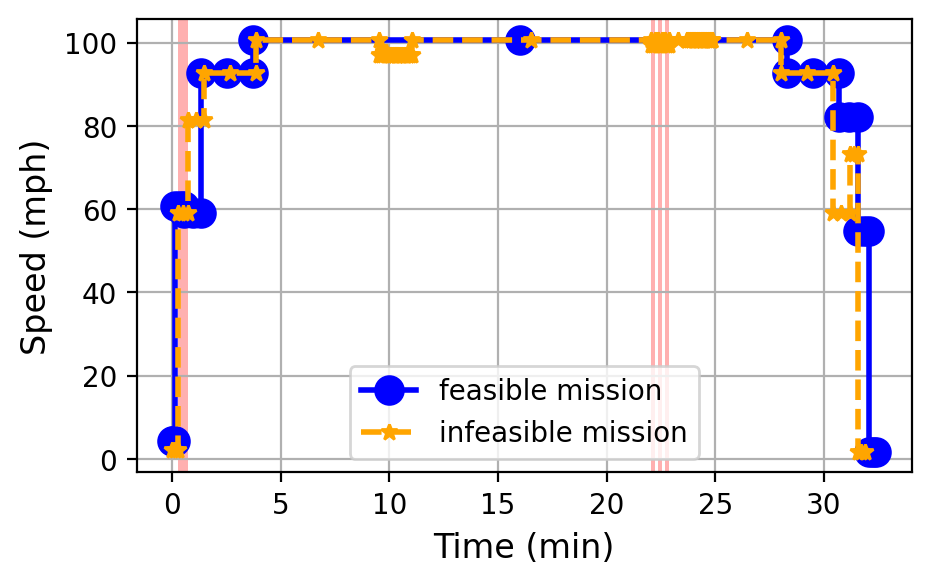}
    \caption{A sample feasible and infeasible airspeed profile. The red vertical bars highlight the segments of the infeasible mission profile where SUAVE eVTOL was unable to execute. }
    \label{fig:speed_profile}
\end{figure}

\begin{figure}
    \centering
    \includegraphics[scale=0.52]{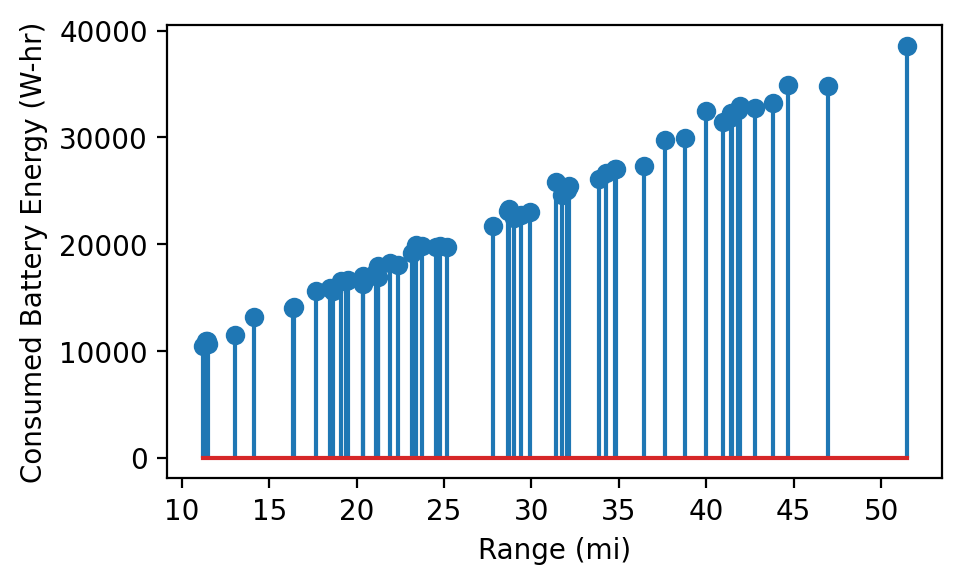}
    \caption{Battery energy consumption of eVTOLs along the mission range. }
    \label{fig:energy_consumption}
\end{figure}

\begin{figure}
    \centering
    \includegraphics[scale=0.52]{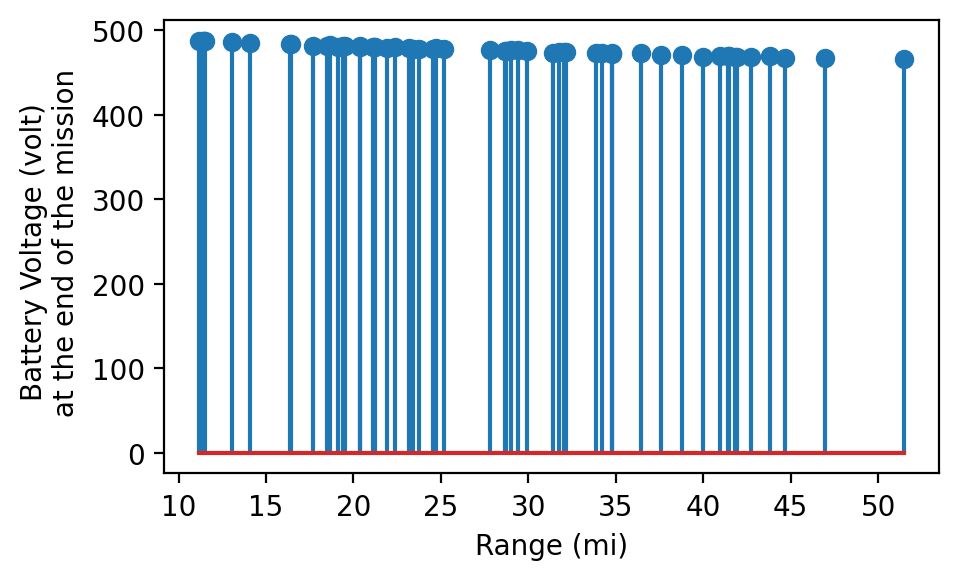}
    \caption{ Battery voltage reading of eVTOLs along the mission range.}
    \label{fig:voltage_consumption}
\end{figure}

\begin{table}[h]
    \centering
    \caption{Baseline mission specification used in the integrated simulation framework.}
    \label{tab:mission_spec}
    \begin{tabular}{c|c|c|c} \hline
         \makecell{Mission \\ segment} & \makecell{Vertical \\speed(ft/min)} & \makecell{Horizontal \\speed(mph)} & \makecell{AGL Ending \\Altitude (ft)} \\ \hline
         Hover Climb                                & 0 to 500   & 0                                   & 50 \\
         Transition + Climb                         & 500        & \makecell{0 to \\$1.2\times v_{stall}$}   & 300 \\
         \makecell{Departure Terminal \\area Procedures} & 0          & $1.2\times v_{stall}$                     & 300 \\
         Accel + Climb                              & 500        & \makecell{$1.2\times v_{stall}$ \\to 110} & 1500 \\
         Cruise                                     & 0          & 110                                 & 1500 \\
         Decel + Descend                            & 500        & \makecell{110 to \\$1.2\times v_{stall}$} & 300 \\
         \makecell{Arrival Terminal \\area Procedures}   & 0 to 500   & $1.2\times v_{stall}$                     & 300 \\
         Transition + Descend                       & 500 to 300 & \makecell{$1.2\times v_{stall}$ \\to 0}   & 50 \\
         Hover + Descend                            & 300 to 0   & 0                                   & 0 \\ \hline
    \end{tabular}
    
\end{table}
From the UTM simulator with the dilation algorithm, we generated 262 full mission profiles. It took 2h 28.21min to execute the performance analysis of the $262$ mission profiles in a workstation with configuration Intel Xeon(R) CPU at 2.2GHz with 88 cores, 128GB RAM, Nvidia Geforce RTX 2080Ti GPU, and Ubuntu OS. In the first step, we conducted feasibility analysis of these 262 mission profiles with the SUAVE tool using the eVTOL shown in Figure \ref{fig:evtol_model}. We found that only 55 mission profiles could be executed by the eVTOL. For the remaining 207 mission profiles, the SUAVE eVTOL model failed to execute at least one segment of the mission profile. \textcolor{black}{Table \ref{tab:comparison_study_with_UTM} shows the performance evaluation comparison between the existing UTM simulator and the proposed simulator. From this comparison study, we can infer that the physical constraints of eVTOL performance can directly impact the applicability of various UTM algorithms.} \\
\indent For further analysis, we show two representative mission profiles, one from the feasible set and another from the infeasible set, in Figure \ref{fig:mission_profile}. The corresponding airspeed profiles are shown in Figure \ref{fig:speed_profile}. \textcolor{black}{From Figure \ref{fig:mission_profile} as indicated by the red vertical bars, the eVTOL was unable to execute the departure terminal area procedure and cruise segments. Since the departure terminal area procedures are generated randomly between a given upper and lower bound, this infeasibility indicates that the considered eVTOL cannot follow any abrupt terminal area procedures. After analyzing the cruise segments of the mission profile using the UTM simulator, we found that the UTM simulator required a certain speed profile to avoid collision, or maintain minimum separation in the airspace. However, the SUAVE eVTOL cannot achieve these speed profiles. These two cases are the indication of sample contingency that will occur in UAM environment. The proposed simulation framework can also be extended to capture other types of contingencies in UAM such as vertiport terminal congestion, adverse weather or emergency landing scenarios.} \\
\indent We continued our analysis using the feasible set of mission profiles to measure other performance metrics. Figure \ref{fig:energy_consumption} shows the consumed battery energy for mission profiles with different ranges. Figure \ref{fig:voltage_consumption} shows the voltage reading of the battery at the end of different mission range values. From Figure \ref{fig:energy_consumption} and \ref{fig:voltage_consumption}, it is clear that the energy consumption of eVTOL is proportional to the mission range while the change in voltage reading is negligible. 

\begin{table}[h]
    \centering
    \caption{Performance evaluation comparison of the proposed simulation framework with baseline UTM simulator }
    \label{tab:comparison_study_with_UTM}
    \begin{tabular}{c|c|c} \hline
        \makecell{Simulation \\framework} & \makecell{Number of feasible\\  mission profiles} &  \makecell{Number of infeasible\\  mission profiles}\\ \hline
        \makecell{Baseline \\ UTM simulator} & 262 & 0\\ \hline
        Proposed simulator & 55 & \makecell{207\\(176 in departure \\ terminal area \\ procedure segment \\ \& 31 in cruise \\segment)}\\ \hline
    \end{tabular}
\end{table}

\begin{figure}[h]
    \centering
    \includegraphics[scale=0.52]{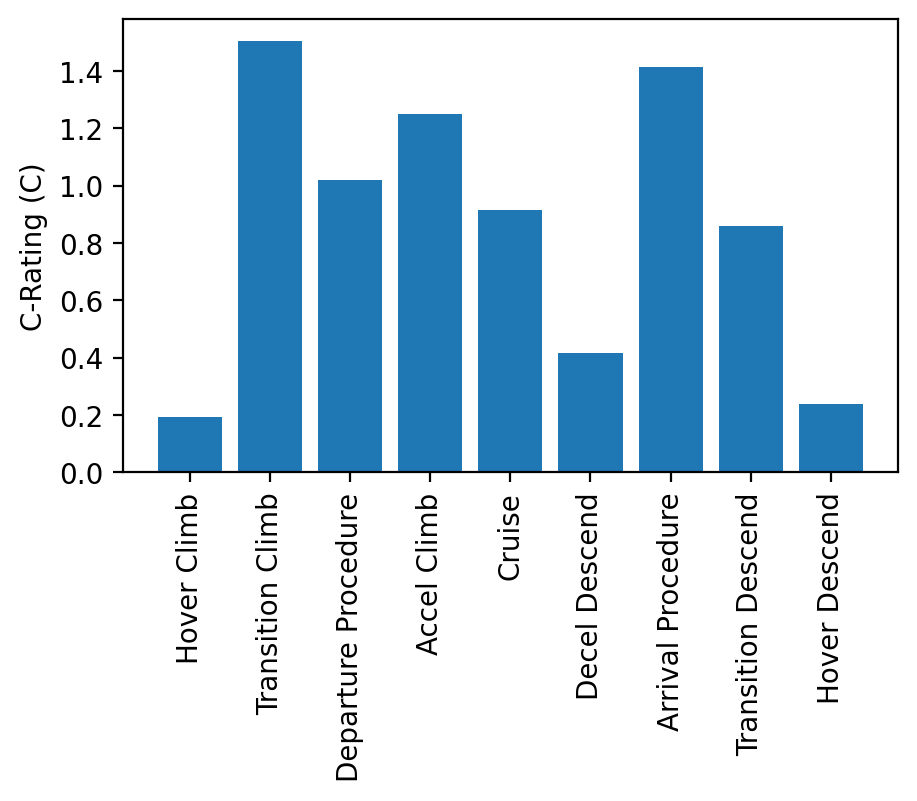}
    \caption{ C-Rating of the eVTOL for different mission segment averaged over the 55 feasible mission profiles.}
    \label{fig:c_rating}
\end{figure}

\begin{figure}[h]
    \centering
    \includegraphics[scale=0.52]{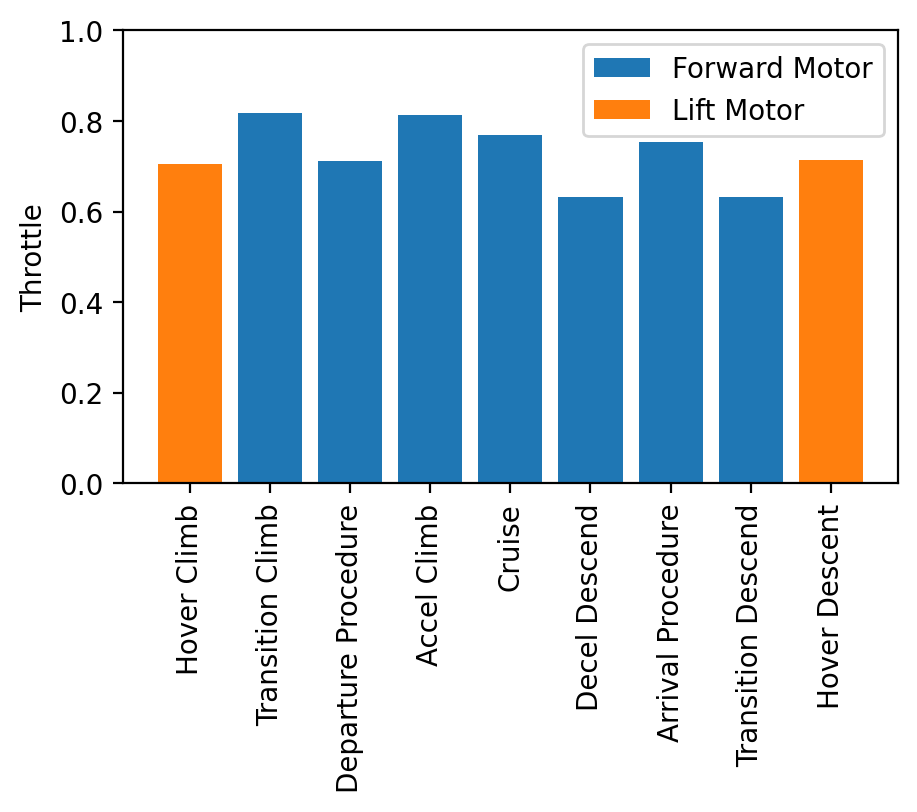}
    \caption{ Throttle from lift \& forward motors of the eVTOL for different mission segment averaged over the 55 feasible mission profiles.}
    \label{fig:throttle}
\end{figure}

In addition, we conducted a C-Rating analysis of the eVTOL's battery and a throttle analysis of the lift and forward motors of the eVTOL for each segment of the mission profile. The C-rating analysis and throttle analysis is displayed in Figures \ref{fig:c_rating} and \ref{fig:throttle}, respectively. From the C-rating analysis, we observe that the eVTOL drives maximum current during the transition climb and arrival procedure. From the throttle analysis, we discover that the lift motors primarily contribute during the vertical motion such as the hover climb and hover descend, while forward motor contributes in all other segments of the mission profile.  

\section{Discussion} \label{discussion}
From the simulation results, the advantages of the proposed simulation framework are summarized below.
\begin{itemize}
    \item The framework provides a more comprehensive and realistic platform for the evaluation of eVTOL's performance in the UAM realm;
    \item It mitigates the limitation of the low-fidelity UTM simulator by capturing and analyzing the infeasible mission profiles using the dilation sub-module;
    \item And the framework can be extended to perform more efficient analysis on selected representative mission profiles for a large-scale of eVTOLs in the UAM network. 
\end{itemize}

\section{Conclusion} \label{conclusion}
In this paper, we presented a new simulation framework to evaluate the performance of eVTOLs utilizing state-of-the-art UTM architectures in the UAM realm. The proposed framework integrates a UTM simulator with an open-source eVTOL design tool to achieve a realistic representation of the UAM environment for evaluation purposes. The developed dilation sub-module provides detailed analysis on the unfeasible missions for the low-fidelity UTM simulator, which considers the trade-off between fidelity and scale. Simulations are performed on the developed framework to demonstrate and analyze the performance of eVTOLs in UAM facilities in terms of different evaluation metrics. From the simulation, the proposed framework not only reflects the details of an eVTOL's performance during different phases of the mission profile, but also captures the occurrence of infeasible mission profiles. Additionally, the proposed framework provides further insights into the specific regions of the infeasible mission profiles to indicate different contingency situations that may occur in future UAM environments.

In the future, the following research directions will be investigated:
\begin{itemize}
    \item The proposed framework currently, employs a model-based performance evaluation procedure for eVOTLs. We will address developing a data-driven performance evaluation tool for the proposed framework to reduce the time complexity.
    \item The current implementation of the proposed framework omits the design of the vertiport and terminal area procedures. We will work to add the vertiport to the proposed framework and build a more realistic simulation interface.
\end{itemize}

\section*{Acknowledgment}

The authors would like to thank Dr. James Paduano from Aurora Flight Sciences for his suggestions during the preparation of the manuscript.

\bibliographystyle{IEEEtran}
\bibliography{ref}

\begin{thebibliography}{10}
\providecommand{\url}[1]{#1}
\csname url@samestyle\endcsname
\providecommand{\newblock}{\relax}
\providecommand{\bibinfo}[2]{#2}
\providecommand{\BIBentrySTDinterwordspacing}{\spaceskip=0pt\relax}
\providecommand{\BIBentryALTinterwordstretchfactor}{4}
\providecommand{\BIBentryALTinterwordspacing}{\spaceskip=\fontdimen2\font plus
\BIBentryALTinterwordstretchfactor\fontdimen3\font minus
  \fontdimen4\font\relax}
\providecommand{\BIBforeignlanguage}[2]{{%
\expandafter\ifx\csname l@#1\endcsname\relax
\typeout{** WARNING: IEEEtran.bst: No hyphenation pattern has been}%
\typeout{** loaded for the language `#1'. Using the pattern for}%
\typeout{** the default language instead.}%
\else
\language=\csname l@#1\endcsname
\fi
#2}}
\providecommand{\BIBdecl}{\relax}
\BIBdecl

\bibitem{sunil2016influence}
E.~Sunil, J.~Hoekstra, J.~Ellerbroek, F.~Bussink, A.~Vidosavljevic,
  D.~Delahaye, and R.~Aalmoes, ``The influence of traffic structure on airspace
  capacity,'' in \emph{ICRAT 2016, 7th International Conference on Research in
  Air Transportation}, 2016.

\bibitem{zhu2016low}
G.~Zhu and P.~Wei, ``Low-altitude uas traffic coordination with dynamic
  geofencing,'' in \emph{16th aiaa aviation technology, integration, and
  operations conference}, 2016, p. 3453.

\bibitem{peinecke2017deconflicting}
N.~Peinecke and A.~Kuenz, ``Deconflicting the urban drone airspace,'' in
  \emph{2017 IEEE/AIAA 36th Digital Avionics Systems Conference (DASC)}.\hskip
  1em plus 0.5em minus 0.4em\relax IEEE, 2017, pp. 1--6.

\bibitem{jang2017concepts}
D.-S. Jang, C.~A. Ippolito, S.~Sankararaman, and V.~Stepanyan, ``Concepts of
  airspace structures and system analysis for uas traffic flows for urban
  areas,'' in \emph{AIAA Information Systems-AIAA Infotech@ Aerospace}, 2017,
  p. 0449.

\bibitem{joulia2016towards}
A.~Joulia, T.~Dubot, and J.~Bedouet, ``Towards a 4d traffic management of small
  uas operating at very low level,'' in \emph{ICAS, 30th Congress of the
  International Council of the Aeronautical Sciences}, 2016.

\bibitem{bulusu2018throughput}
V.~Bulusu, R.~Sengupta, E.~R. Mueller, and M.~Xue, ``A throughput based
  capacity metric for low-altitude airspace,'' in \emph{2018 Aviation
  Technology, Integration, and Operations Conference}, 2018, p. 3032.

\bibitem{ramee2021development}
C.~Ramee and D.~N. Mavris, ``Development of a framework to compare low-altitude
  unmanned air traffic management systems,'' in \emph{AIAA Scitech 2021 Forum},
  2021, p. 0812.

\bibitem{lukaczyk2015suave}
T.~W. Lukaczyk, A.~D. Wendorff, M.~Colonno, T.~D. Economon, J.~J. Alonso, T.~H.
  Orra, and C.~Ilario, ``Suave: an open-source environment for multi-fidelity
  conceptual vehicle design,'' in \emph{16th AIAA/ISSMO Multidisciplinary
  Analysis and Optimization Conference}, 2015, p. 3087.

\bibitem{clarke2019strategies}
M.~Clarke, J.~Smart, E.~M. Botero, W.~Maier, and J.~J. Alonso, ``Strategies for
  posing a well-defined problem for urban air mobility vehicles,'' in
  \emph{AIAA Scitech 2019 Forum}, 2019, p. 0818.

\bibitem{chambers2011reforms}
C.~Chambers, ``The reforms: a political safe haven or political suicide--is the
  labour bubble bursting?'' \emph{Journal of Financial Regulation and
  Compliance}, 2011.

\bibitem{effectiveness2018}
P.~Sachs, C.~Dienes, E.~Dienes, and M.~Egorov, ``Effectiveness of preflight
  deconfliction in highdensity uas operations,'' in \emph{Tech rep, Altiscope},
  2018.

\bibitem{sedov2018centralized}
L.~Sedov and V.~Polishchuk, ``Centralized and distributed utm in layered
  airspace,'' \emph{8th ICRAT}, 2018.

\bibitem{eby1994self}
M.~S. Eby, ``A self-organizational approach for resolving air traffic
  conflicts.'' \emph{Lincoln Laboratory Journal}, 1994.

\bibitem{hoekstra2016bluesky}
J.~M. Hoekstra and J.~Ellerbroek, ``Bluesky atc simulator project: an open data
  and open source approach,'' in \emph{Proceedings of the 7th International
  Conference on Research in Air Transportation}, vol. 131.\hskip 1em plus 0.5em
  minus 0.4em\relax FAA/Eurocontrol USA/Europe, 2016, p. 132.

\bibitem{UBAR_mission_spec}
\BIBentryALTinterwordspacing
Uber, ``Uber air vehicle requirements and missions,'' 2019. [Online].
  Available:
  \url{"https://s3.amazonaws.com/uber-static/elevate/Summary+Mission+and+Requirements.pdf"}
\BIBentrySTDinterwordspacing

\bibitem{SUAVEGit}
\BIBentryALTinterwordspacing
A.~Wendorff, A.~Variyar, C.~Ilario, E.~Botero, F.~Capristan, J.~Smart,
  J.~Alonso, L.~Kulik, M.~Clarke, M.~Colonno, M.~Kruger, J.~M. Vegh,
  P.~Goncalves, R.~Erhard, R.~Fenrich, T.~Orra, T.~St.~Francis, T.~MacDonald,
  T.~Momose, T.~Economon, T.~Lukaczyk, and W.~Maier, ``Suave: An aerospace
  vehicle environment for designing future aircraft,'' 2020. [Online].
  Available: \url{https://github.com/suavecode/SUAVE}
\BIBentrySTDinterwordspacing

\bibitem{SUAVE2017}
\BIBentryALTinterwordspacing
T.~MacDonald, M.~Clarke, E.~M. Botero, J.~M. Vegh, and J.~J. Alonso,
  \emph{SUAVE: An Open-Source Environment Enabling Multi-Fidelity Vehicle
  Optimization}. [Online]. Available:
  \url{https://arc.aiaa.org/doi/abs/10.2514/6.2017-4437}
\BIBentrySTDinterwordspacing

\bibitem{holden2016fast}
J.~Holden and N.~Goel, ``Fast-forwarding to a future of on-demand urban air
  transportation,'' \emph{San Francisco, CA}, 2016.

\end{thebibliography}

\end{document}